\documentclass[sigconf]{acmart}

\copyrightyear{2024}
\acmYear{2024}
\setcopyright{licensedothergov}\acmConference[SIGIR '24]{Proceedings of the 47th International ACM SIGIR Conference on Research and Development in Information Retrieval}{July 14--18, 2024}{Washington, DC, USA}
\acmBooktitle{Proceedings of the 47th International ACM SIGIR Conference on Research and Development in Information Retrieval (SIGIR '24), July 14--18, 2024, Washington, DC, USA}
\acmDOI{10.1145/3626772.3657901}
\acmISBN{979-8-4007-0431-4/24/07}

\usepackage{graphicx}
\usepackage{multirow}
\usepackage{url}
\usepackage{amsmath}
\usepackage[utf8]{inputenc}
\usepackage{balance}

\begin{document}

\title{From Text to Context: An Entailment Approach for News Stakeholder Classification}


\author{Alapan Kuila}
\orcid{0009-0006-4168-1190}
\affiliation{%
  \institution{IIT KHARAGPUR}
  \city{Kharagpur}
  \country{India}
}
\email{alapan.cse@gmail.com}

\author{Sudeshna Sarkar}
\orcid{0000-0003-3439-4282}
\affiliation{%
  \institution{IIT KHARAGPUR}
  \city{Kharagpur}
  \country{India}}
\email{sudeshna@cse.iitkgp.ac.in}


\renewcommand{\shortauthors}{Alapan Kuila and Sudeshna Sarkar}
\begin{abstract}
Navigating the complex landscape of news articles involves understanding the various actors or entities involved, referred to as news stakeholders. These stakeholders, ranging from policymakers to opposition figures, citizens, and more, play pivotal roles in shaping news narratives. Recognizing their stakeholder types, reflecting their roles, political alignments, social standing, and more, is paramount for a nuanced comprehension of news content. Despite existing works focusing on salient entity extraction, coverage variations, and political affiliations through social media data, the automated detection of stakeholder roles within news content remains an underexplored domain. In this paper, we bridge this gap by introducing an effective approach to classify stakeholder types in news articles. Our method involves transforming the stakeholder classification problem into a natural language inference task, utilizing contextual information from news articles and external knowledge to enhance the accuracy of stakeholder type detection. Moreover, our proposed model showcases efficacy in zero-shot settings, further extending its applicability to diverse news contexts.
\end{abstract}






\maketitle

\section{Introduction}
In the intricate world of news reporting, especially when dealing with various government policies on socio-political and economic fronts, mass media frequently spotlights key players and entities directly or indirectly engaged in these issues. These entities, including policy-makers (like government officials), political opponents (representing opposition political parties), consumers (ordinary citizens), and civic society organizations, take center stage as opinion holders, significantly steering the overall tone and direction of news content. Referred to as \textit{Stakeholders} in our work, these influential entities hold considerable sway over public discourse. The significance of our task lies in the ongoing competition among stakeholders for increased media coverage, a competition driven by the desire to heighten visibility among news consumers~\cite{eberl2017one}. Moreover, research in journalism and digital media has delved into the influence of the political inclinations or social stature of these key stakeholders, providing insights into the inherent ideologies and political leanings of publishers~\cite{wilkerson2017large}. Notably, the politicization of COVID-pandemic-related news, where politicians receive more coverage than scientists and researchers~\cite{hart2020politicization}, exemplifies the urgency to address the task at hand—grouping these essential entities within news articles into relevant stakeholder classes~\cite{10.1145/3388142.3388149}. This involves identifying the stakeholder type of individual entities, thereby facilitating the recognition of actors sharing similar political views or socio-economic backgrounds—a fundamental challenge in the domain of computational journalism~\cite{sharma2020ideology}.

In the domain of recognizing stakeholder classes, understanding the roles of key actors~\cite{dunietz2014new, wu2020all} in news topics and their involvement in particular issues becomes paramount. Existing research in this field reveals certain research gaps that our work aims to address. Previous studies, particularly in the financial domain~\cite{repke2021extraction, ponza2021contextualizing} have focused on identifying salient entities~\cite{10.1145/2505515.2505602} but primarily relied on extracting target company names from news articles~\cite{Grnberg2021ExtractingSN, gamon2013understanding}. While some efforts have been made to extract influential entities and determine their political affiliations and roles using various online resources~\cite{sen2019attempt}, this approach is cumbersome and often relies on incomplete knowledge bases. Notably, each news topic pertaining to a specific government policy involves a distinct set of potential stakeholders~\cite{sen2018leveraging}. While common stakeholders like the government, opposition, and citizens are prevalent across various policies, there exist topic-specific stakeholders depending on the nature of the government policy, such as the Banking sector, private sectors (in Economic policies), and foreign nations and their leaders (in Foreign policies). The heterogeneous nature of the news domain and the domain dependency of stakeholder classes render the task challenging.

Comparable works, such as those focusing on political preference or ideology detection of salient actors, often rely on social media posts and metadata from social media accounts~\cite{tang2010community, benton2016learning, pan2019social, budhiraja2021american, haq2023twitter}. However, the brief nature of social media posts and the specific structural configuration of social network sites facilitate better representation learning of stakeholder entities, aiding in more accurate political perspective detection~\cite{gu2021exploiting}. In contrast, news articles are descriptive and contain textual clues crucial for entity representation learning spread throughout the article. Occasionally, a specific news document may lack sufficient information for effective stakeholder-type classification. Our work aims to bridge these gaps by proposing an efficient approach that considers the heterogeneous nature of news articles and mitigates the challenges posed by the varied domain dependencies of stakeholder classes.

In this paper, we tackle the challenge of classifying stakeholder types for salient entities mentioned in news articles detailing Indian government policies. Recognizing the need for addressing new stakeholder classes within the heterogeneous news domain, we frame the classification problem as an entailment-based natural language inference task~\cite{yin2019benchmarking}. Inspired by ~\cite{song2014dataless, chen2015dataless, kabongo2023zero}, we develop an entailment-based zero-shot classifier for stakeholder type detection, addressing the scarcity of labeled datasets~\cite{chang2008importance} for topic-specific stakeholder classes. Our approach leverages both cross-document context as textual knowledge and excerpts from Wikipedia pages as global information for effective entity feature representation. We construct a weakly supervised textual entailment dataset, incorporating stakeholder entity descriptions and label information. Through training our proposed model\footnote{Link to code and dataset: \url{https://github.com/alapanju/NewsStake}} on the entailment task, it adeptly classifies salient entity mentions in news articles into unseen stakeholder classes with a high degree of success.

\section{Task Definition}

The news stakeholder classification task involves labeling prominent entities in news articles based on factors like their roles, social standing, political alignment, and geographic location. Our framework aims to determine each entity's stakeholder class by analyzing its description within news articles, offering insights into their perspectives on various news topics. We denote an entity mention as $x = <e, M>$, where $e$ represents the entity phrase and $M$ comprises snippets from multiple news articles containing $e$ or its coreference mentions. We enhance $M$ with information from external knowledge sources (e.g., Wikipedia) to create the aggregated entity description $M' = w \oplus M$. Our objective is to design a topic-agnostic stakeholder classifier $f$ that uses $M'$ to accurately detect the true stakeholder classes ($\in S$) of all prominent entities mentioned in heterogeneous news articles. Formally, $f: e \times M' \rightarrow S$. In subsequent sections, we outline our approach to formalizing the classification task into an entailment problem, detail the method for extracting cross-document entity representations, model description and analyze the results.

\begin{table}[!htp]
\center\fontsize{6}{10}\selectfont
\caption{Transform the classification dataset into NLI dataset. The output label 1 if the premise entails the hypothesis and 0 otherwise. $prompt(e,S)$ denotes the prompt template mentioning both entity phrase and the stakeholder label.}
\label{table:transform}
\begin{tabular}{|ccccc|}
\hline
\multicolumn{5}{|c|}{Topic: T}                                                                                                                                                                                                                                                                                                   \\ \hline
\multicolumn{5}{|c|}{Stakeholder Types: S1, S2}                                                                                                                                                                                                                                                                                   \\ \hline
\multicolumn{3}{|c||}{\begin{tabular}[c]{@{}c@{}}Classification Dataset\end{tabular}}                                                                                                                                                & \multicolumn{2}{c|}{\begin{tabular}[c]{@{}c@{}}Formatting into  NLI task\end{tabular}} \\ \hline
\multicolumn{1}{|c|}{\begin{tabular}[c]{@{}c@{}}Entity Phrase\end{tabular}} & \multicolumn{1}{c|}{Description}                                                                           & \multicolumn{1}{c||}{Label}               & \multicolumn{1}{c|}{\begin{tabular}[c]{@{}c@{}}Input Sequence\end{tabular}}   & Label  \\ \hline
\multicolumn{1}{|c|}{\multirow{2}{*}{e}}                                      & \multicolumn{1}{c|}{\multirow{2}{*}{\begin{tabular}[c]{@{}c@{}}M\\ (Textual description)\end{tabular}}} & \multicolumn{1}{c||}{\multirow{2}{*}{S1}} & \multicolumn{1}{c|}{M [SEP] $prompt(e, S1)$}                                    & 1      \\ \cline{4-5} 
\multicolumn{1}{|c|}{}                                                        & \multicolumn{1}{c|}{}                                                                                      & \multicolumn{1}{c||}{}                    & \multicolumn{1}{c|}{M [SEP] $prompt(e, S2)$}                                    & 0      \\ \hline \hline
\multicolumn{5}{|l|}{\begin{tabular}[c]{@{}l@{}}Premise: entity description (M)\\ Hypothesis: Sentence with label information ($prompt(e, S)$)\\ $prompt(e, S)$: The entity \{e\} belongs to the stakeholder group of \{S\}\\ {[}SEP{]}: Sentence delimiter\end{tabular}}                                                        \\ \hline

\end{tabular}

\end{table}

\section{Our Approach}

\subsection{Formalizing Classification as Entailment}

\begin{figure*}[!htp]
  \centering
  \includegraphics[width=0.80\textwidth]{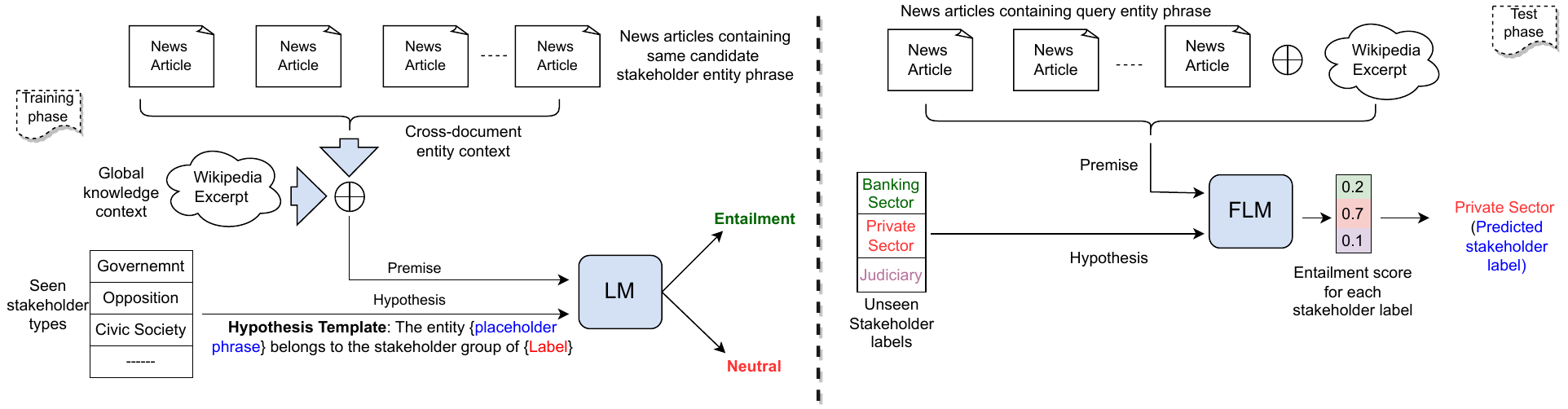}
  \caption{Illustration of the entailment approach for zero-shot stakeholder classification. The left-hand side depicts how the model is trained on entailment task, and on the right-hand side, we demonstrate how the fine-tuned model predicts new stakeholder classes for the query entity. }
  \label{fig:framework}
\end{figure*}

Addressing the challenge of a lack of properly labeled datasets for news stakeholder classification, we reframe the problem as a natural language inference (NLI) task. This strategic shift enables us to fine-tune a domain-agnostic zero-shot classifier, circumventing the need for topic-specific labeled datasets. Initially, we annotate entity mentions with stakeholder labels based on their descriptions from multiple news documents and Wikipedia pages. These annotations form the basis of our transformed NLI dataset, where each instance comprises a premise (entity description) and a hypothesis (stakeholder label-embedded prompt) as detailed in Table \ref{table:transform}. During training, the model learns to predict whether the premise entails the corresponding hypothesis. In testing, the fine-tuned model utilizes entailment scores to classify query entities into unseen classes as illustrated in Figure \ref{fig:framework}, thus effectively addressing the challenge of a lack of labeled data for specific stakeholder classes.

\subsection{Entity Representation}
\label{Entity Representation}

In this section, we delineate the procedure for generating entity descriptions for the stakeholder entity phrases.

\paragraph{Entity Identification}
Initially, we utilize Spacy's~\footnote{\url{https://spacy.io/api/entityrecognizer}} entity recognizer to identify all entity mentions within news articles. This tool enables the extraction of entity phrases along with their associated entity types. However, our focus is narrowed to specific entity types with potential stakeholder significance, including \textit{Person}, \textit{Geopolitical-entity}, and \textit{Organization}. Additionally, we impose a constraint that entities must exhibit salience within the document context to qualify as valid stakeholders. To determine saliency, we consider entities referenced multiple times within the document, ensuring their significance and relevance as potential stakeholders.

\paragraph{WD-Entity Context} 
We identify relevant sentences within news articles containing the target entity or its coreference mentions to form the Within Document (WD) entity description. Leveraging the LINGMESS coreference model~\cite{otmazgin-etal-2023-lingmess}, we ensure comprehensive identification of all entity coreference chains within the document.

\paragraph{CD-Entity Context}

To address limited information in single documents, we extend our analysis across multiple documents, resolving cross-document entity references crucial for accurate stakeholder prediction. Using string matching and phonetic measures like Jaro-Winkler similarity\footnote{Experimentally, threshold values between 0.8 to 0.9 yield optimal results for effective cross-document entity resolution.}, Levenstein distance, and substring matching, we identify relevant cross-document coreference mentions. Aggregating entity context from individual articles, we form the Cross-Document (CD) entity description, providing a comprehensive stakeholder representation across documents.

\paragraph{Background Knowledge as Entity Context}

We enhance entity descriptions by integrating external domain knowledge from relevant Wikipedia pages. Initially, we link target entity phrases to corresponding Wikipedia pages and extract introductory sentences for additional information. Leveraging the Wikipedia Python library\footnote{\url{https://pypi.org/project/wikipedia/}}, we access and parse Wikipedia data, overcoming challenges in identifying correct pages for some stakeholder phrases. To address this, we manually retrieve India-related Wikipedia pages, enriching our dataset with substantial India-specific textual content. This augmentation significantly enhances the context and depth of our stakeholder descriptions.

\section{Data} 
In this section, we will depict the dataset creation procedure along with the dataset description.

\begin{table*}[!ht]
\center\fontsize{7}{8}\selectfont
\caption{The Stakeholder considered in each news topic. Topic-specific stakeholders are indicated in bold fonts.}
\label{table:stakeholder_types}
\begin{tabular}{c|l}
\hline
Topic          & \multicolumn{1}{c}{Stakeholders}                                                                                                                              \\ \hline
Agriculture Act     & \begin{tabular}[c]{@{}l@{}}Government,  Opposition, Citizen/Activists, Bureaucrat, \textbf{Farmers}, \textbf{International-figure}\end{tabular}                                  \\ \hline
Demonetization & \begin{tabular}[c]{@{}l@{}}Government, Opposition, Citizen/Activists, Bureaucrat, \textbf{Banking Sector}, \textbf{Private Companies}\end{tabular}         \\ \hline
CAB Bill       & \begin{tabular}[c]{@{}l@{}}Government,  Opposition, Citizen/Activist, Bureaucrat, \textbf{International-figure}\end{tabular}     \\ \hline
COVID Control & \begin{tabular}[c]{@{}l@{}}Government, Opposition, Citizen/Activist\end{tabular}, Bureaucrat,  \textbf{Scientist/Researchers}, \textbf{International-figure} \\ \hline

Article 370 & \begin{tabular}[c]{@{}l@{}}Government, Opposition, Citizen/Activist\end{tabular}, Bureaucrat, \textbf{International-figure, Judiciary, Kashmiri people} \\ \hline
\end{tabular}
\end{table*}

\paragraph{News Domain} All our experiments are based on the news articles on five Indian Govt policies:
1) Agriculture Act (2020)\footnote{\url{https://en.wikipedia.org/wiki/2020_Indian_agriculture_acts}}, 2) Demonetization\footnote{\url{https://en.wikipedia.org/wiki/2016_Indian_banknote_demonetisation}}, 3) Citizenship Amendment Bill (CAB)\footnote{\url{https://en.wikipedia.org/wiki/Citizenship_(Amendment)_Act,_2019}}, 4) COVID pandemic management\footnote{\url{https://en.wikipedia.org/wiki/COVID-19_pandemic_in_India}} and 5)Abrogation of Article 370\footnote{\url{https://en.wikipedia.org/wiki/Revocation_of_the_special_status_of_Jammu_and_Kashmir}}.


    


\paragraph{Stakeholder type selection} 
After consulting with domain experts (PhD scholars) from political science and social science backgrounds, we finalized a set of stakeholder groups for the aforementioned news topics. Given the multi-party system in India, various potential stakeholder groups exist, and we identified some of the prominent ones for our experiment. Certain stakeholders, such as political parties (both ruling and opposition), elected government officials, and bureaucrats, are common across all political news topics. Moreover, the stakeholder type "Citizen \& Activist" plays a significant role in societal discourse, with their voices carrying substantial influence. Recognizing the pivotal role of media agencies in political news coverage, they are also considered as potential stakeholders. However, certain stakeholders are relevant only to specific news topics. The complete list of stakeholder groups considered in our experiment is provided in Table~\ref{table:stakeholder_types}.

\begin{table}[!ht]
\center\fontsize{7}{10}\selectfont
\caption{Statistics of labelled dataset in each news topic}
\label{dataset_stat}
\begin{tabular}{c|c|c}
\hline
\textbf{News Topic}                                                & \textbf{\begin{tabular}[c]{@{}c@{}}Number of \\ Stakeholder Labels\end{tabular}} & \textbf{\begin{tabular}[c]{@{}c@{}}Number\\ of Instances\end{tabular}} \\ \hline
\textbf{Agriculture Act} & 6                                                                    & 302                                                                    \\ 
\textbf{Controlling COVID}   & 6                                                                    & 351                                                                    \\ 
\textbf{CAB}                                                       & 5                                                                    & 252                                                                    \\ \hline
\textbf{Demonetization}                                            & 6                                                                    & 250                                                                    \\ 
\textbf{Article 370}                                               & 7                                                                    & 253                                                                    \\ \hline
\end{tabular}
\end{table}

\paragraph{Article Collection}

We gather topic-oriented news articles from \textit{GDELT}\footnote{\url{https://blog.gdeltproject.org/}} and \textit{EventRegistry}\footnote{\url{https://github.com/EventRegistry/event-registry-python}}. \textit{GDELT} provides a dataset of geo-located events reported in news articles worldwide, while \textit{EventRegistry} offers access to news data through its API. We extract the actual content from URLs provided by \textit{GDELT GKG table}\footnote{\url{https://blog.gdeltproject.org/}} using scraping tools. \textit{EventRegistry}'s Python package facilitates filtering articles based on various parameters. After extraction, we use a bag-of-words approach and semi-supervised LDA~\cite{wang2012semi} to identify topic-specific news articles.

\paragraph{Annotation Procedure}

In the absence of existing annotated data fitting our research scope, we constructed a detailed annotation guide providing thorough descriptions of stakeholder classes and their interrelations. Two domain experts, both PhD scholars in India-specific political news, annotated stakeholder types using entity representations from Section~\ref{Entity Representation}. Any labeling uncertainties were resolved through discussion, resulting in a customized dataset for stakeholder classification.

\paragraph{Data Statistics}
In Table~\ref{dataset_stat}, we report the statistics of the annotated dataset. We use news topics: \textit{Agriculture Act, COVID Control} and \textit{CAB} for training our model. For evaluation purpose we use new topics \textit{demonetization} and \textit{Article 370}. The number of labels and label specific training and development and test sets of our dataset used for our experiment is reported in Table~\ref{train_data}.

\begin{table}[!ht]
\center\fontsize{7}{10}\selectfont
\caption{Number of train, development and test sets of our dataset. Here \textit{Dev} indicates development set. $Test_{seen}$ indicates test set containing labels present in the training set. Labels in $Test_{unseen}$ are used to evaluate the model performance in zero-shot settings.}
\label{train_data}
\begin{tabular}{c|c|c}
\hline
\textbf{Dataset} & \textbf{Number of labels} & \textbf{Number of examples} \\ \hline
Train            & 7                         & 674                         \\ \hline
Dev              & 7                         & 231                         \\ \hline
$Test_{seen}$    & 5                         & 225                         \\ \hline
$Test_{unseen}$  & 4                         & 278                         \\ \hline
\end{tabular}
\end{table}

\section{Experiments}

Our aim is to develop a model that receives an entity description as input and produces entailment scores for each candidate stakeholder label. The ultimate prediction involves selecting the highest-scoring class in a single-label classification setup or the top-K classes surpassing a specified threshold in a multi-label scenario. In our experimental setup, we assign a single stakeholder label to each candidate entity, providing a streamlined approach to stakeholder-type prediction.

\subsection{Model Description}
For addressing the stakeholder classification problem, we utilize two distinct model architectures: 1) an encoder-only model and 2) an encoder-decoder model. In the encoder-only model, we fine-tune the RoBERTa model~\cite{zhuang-etal-2021-robustly}, comprising 355M parameters. Alternatively, for the encoder-decoder model, we fine-tune the BART model~\cite{Lewis2019BARTDS}, equipped with 400M parameters. 

\subsection{Results and Discussions}
The performance of our two models in stakeholder type classification is showcased in Table~\ref{table_result}. The "Known Labels" column presents the results on the $Test_{seen}$ dataset, while the "Unknown Labels" column reports the performance on the $Test_{unseen}$ dataset, representing the model's efficacy in a zero-shot setting. Table~\ref{table_result} shows that both models exhibit comparable performance when classifying stakeholder labels in supervised settings. However, in zero-shot settings, the RoBERTa model outperforms the BART model by a slight margin. These results form the basis for our discussion on the effectiveness of the proposed models in handling both seen and unseen stakeholder labels.

\begin{table}[!ht]
\caption{Performance of our two models in classifying stakeholder types in supervised and zero-shot settings.}
\label{table_result}
\center\fontsize{7}{8}\selectfont
\begin{tabular}{c|ccc||ccc}
\hline
\multirow{2}{*}{\textbf{Model}}                          & \multicolumn{3}{c|}{\textbf{\begin{tabular}[c]{@{}c@{}}Known\\  Labels\end{tabular}}}                         & \multicolumn{3}{c}{\textbf{\begin{tabular}[c]{@{}c@{}}Unknown\\  Labels\end{tabular}}}                       \\ \cline{2-7} 
                                                         & \multicolumn{1}{c|}{P}     & \multicolumn{1}{c|}{R}     & \begin{tabular}[c]{@{}c@{}}F1-\\ Score\end{tabular} & \multicolumn{1}{c|}{P}     & \multicolumn{1}{c|}{R}     & \begin{tabular}[c]{@{}c@{}}F1\\ -score\end{tabular} \\ \hline
\begin{tabular}[c]{@{}c@{}}RoBERTa\\ -large\end{tabular} & \multicolumn{1}{c|}{86.71} & \multicolumn{1}{c|}{95.77} & 90.46                                               & \multicolumn{1}{c|}{77.56} & \multicolumn{1}{c|}{81.43} & 79.44                                               \\ \hline
\begin{tabular}[c]{@{}c@{}}BART\\ -large\end{tabular}    & \multicolumn{1}{c|}{84.93} & \multicolumn{1}{c|}{95.30} & 89.82                                               & \multicolumn{1}{c|}{74.68} & \multicolumn{1}{c|}{79.11} & 76.83                                               \\ \hline
\end{tabular}

\end{table}

\subsection{Model Robustness}

In this section, we explore the influence of hypothesis prompt-templates on model performance in a zero-shot setting. We assess this impact by employing two semantically equivalent hypothesis templates with distinct tokens and evaluating the resulting model performance. Furthermore, we compare our findings with the widely used NLI-based zero-shot classifier, \textit{bart-large-mnli} from the \textit{HuggingFace Hub}.

\begin{figure}[!htp]
  \centering
  \includegraphics[width=0.8\linewidth]{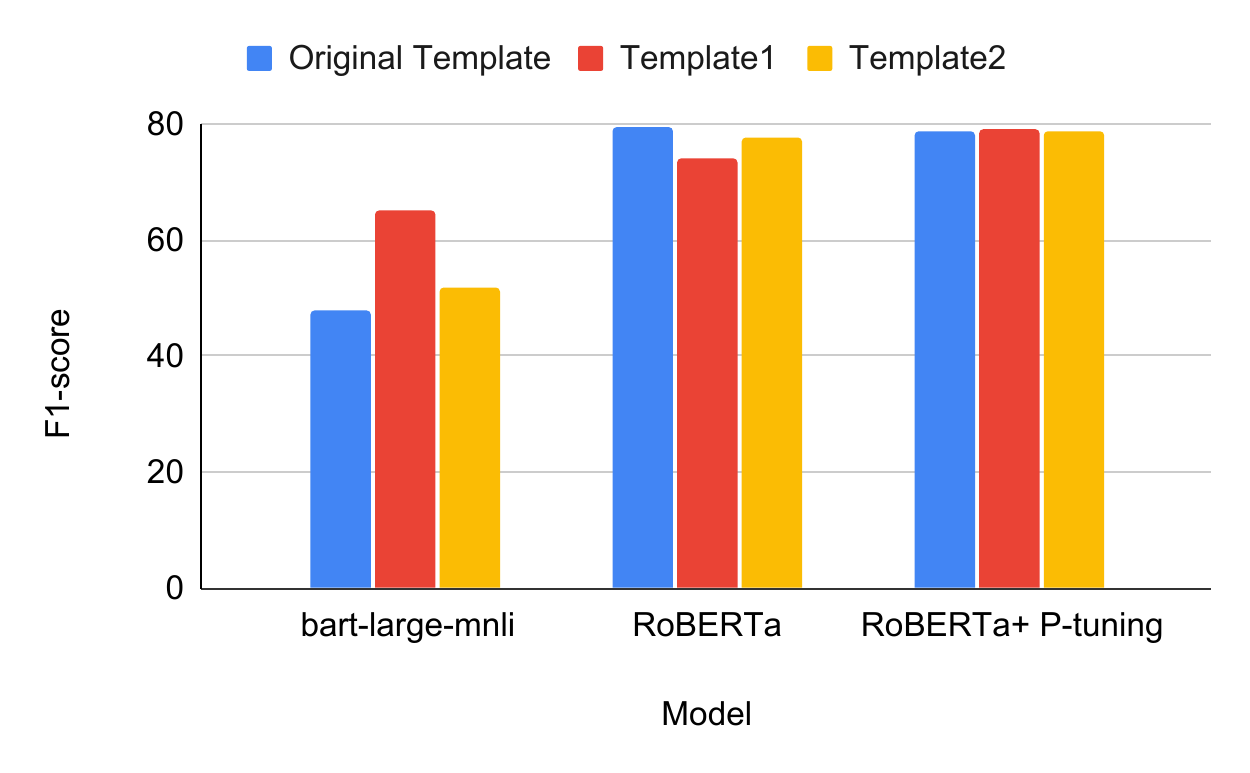}
  \caption{Zero-shot classification Performance of RoBERTa, RoBERTa+P-tuning and bart-large-mnli model (from Facebook) on different hypothesis templates.  Here,  \textit{Template1}: \textbf{The entity \textit{\{placeholder phrase\}} is \textit{\{placeholder label\}}}; \textit{Template2}: \textbf{The entity \textit{\{placeholder phrase\}} is of stakeholder type \textit{\{placeholder label\}}}; and       \textit{Original Template} refers to the template mentioned in the Table 1. }
  \label{model_robust}
\end{figure}

Figure~\ref{model_robust} illustrates that the performance of the \textit{bart-large-mnli} model exhibits instability. Our proposed model's F1-score also varies by $2\%$ and $5\%$ when using different prompt templates. To address this issue, we employ P-tuning~\cite{liu2021gpt}, which utilizes trainable continuous prompt embeddings in conjunction with discrete prompts and trains the RoBERTa model. P-tuning enhances the model's robustness against changes in the hypothesis template.

\section{Conclusion and Future Work}

In this paper, we propose a novel approach for stakeholder classification in news articles, leveraging natural language inference and zero-shot classifiers. Our method offers valuable insights into news narratives, demonstrating effectiveness in both seen and unseen scenarios. Additionally, we explore methods to design robust and stable zero-shot classifiers. Moving forward, we aim to enhance zero-shot model performance, predict finer stakeholder labels, and uncover news bias through stakeholder coverage analysis.

\bibliographystyle{ACM-Reference-Format}
\balance
\bibliography{sample-authordraft}

\appendix









\end{document}